%% file: main.tex
\documentclass[10pt,twocolumn,letterpaper,pagenumbers]{article}

\usepackage[pagenumbers]{wacv} % To force page numbers, e.g. for an arXiv version

\usepackage{graphicx}
\usepackage{amsmath}
\usepackage{amssymb}
\usepackage{booktabs}
\usepackage{multirow}
\usepackage{multicol}
\usepackage{ulem}
\usepackage{stfloats}
\usepackage{enumitem}
\usepackage{subcaption}
\usepackage{mwe}
\usepackage{stfloats}
\usepackage{subcaption}
\usepackage{caption}
\usepackage{cuted}    % provides the strip environment
\usepackage{capt-of}  % provides \captionof
\usepackage{colortbl}

\definecolor{lightgreen}{rgb}{0.7176, 0.8902, 0.7216}
\definecolor{lightyellow}{rgb}{0.9843, 0.9765, 0.7020}


\definecolor{wacvblue}{rgb}{0.21,0.49,0.74}
\usepackage[pagebackref,breaklinks,colorlinks,allcolors=wacvblue]{hyperref}

\def\wacvPaperID{426} % *** Enter the WACV Paper ID here
\def\confName{WACV}
\def\confYear{2026}

\title{HYDRA: HYbrid knowledge Distillation and spectral Reconstruction Algorithm for high channel hyperspectral camera applications}

\author{Christopher Thirgood\\
University of Surrey\\
Surrey, UK\\
{\tt\small c.thirgood@surrey.ac.uk}
\and
Oscar Mendez\\
University of Surrey\\
Surrey, UK\\
{\tt\small o.mendez@surrey.ac.uk}
\and
Erin Chao Ling\\
University of Surrey\\
Surrey, UK\\
{\tt\small chao.ling@surrey.ac.uk}
\and
Jon Storey\\
i3D Robotics\\
Kent, UK\\
{\tt\small Jstorey@i3drobotics.com}
\and
Simon Hadfield\\
University of Surrey\\
Surrey, UK\\
{\tt\small s.hadfield@surrey.ac.uk}
}

\begin{document}

\maketitle

% \begin{figure*}[!ht]
% \vspace{-5mm}
%     \centering
%     \begin{subfigure}[ht]{0.33\textwidth}
%         \includegraphics[width=\textwidth]{images/images/Figure_1_a_lin_grided_cropped.pdf}
%         \caption{NTIRE-2022 Dataset}
%         \label{fig:image1}
%     \end{subfigure}
%     % \hfill % This will add some space between the two subfigures
%     \begin{subfigure}[ht]{0.33\textwidth}
%         \includegraphics[width=\columnwidth]{images/images/Figure_1_b_lin_grid_cropped.pdf}
%         \caption{HySpecNet-11k Dataset}
%         \label{fig:image2}
%     \end{subfigure}
%     % \hfill % This will add some space between the two subfigures
%     \begin{subfigure}[ht]{0.33\textwidth}
%         \includegraphics[width=\columnwidth]{images/images/Figure_1_libhsi_grid_lin_cropped.pdf}
%         \caption{LIB-HSI Datasets}
%         \label{fig:image2}
%     \end{subfigure}
%     \vspace{-1mm}
% 	\caption{
%  HYDRA excels in spectral reconstruction across various channel depths, balancing computational efficiency (FLOPS) with high performance (PSNR). Circle size indicates memory cost (Params). Datasets include 31-channel (a), 202-channel (b), and 204-channel (c).
%  }
% 	\label{fi:tParametersOverview}
%  \vspace{-6.0mm} 
% \end{figure*}

\begin{strip}
  \centering
  %–– your three side-by-side graphics ––%
  \includegraphics[width=0.32\textwidth]{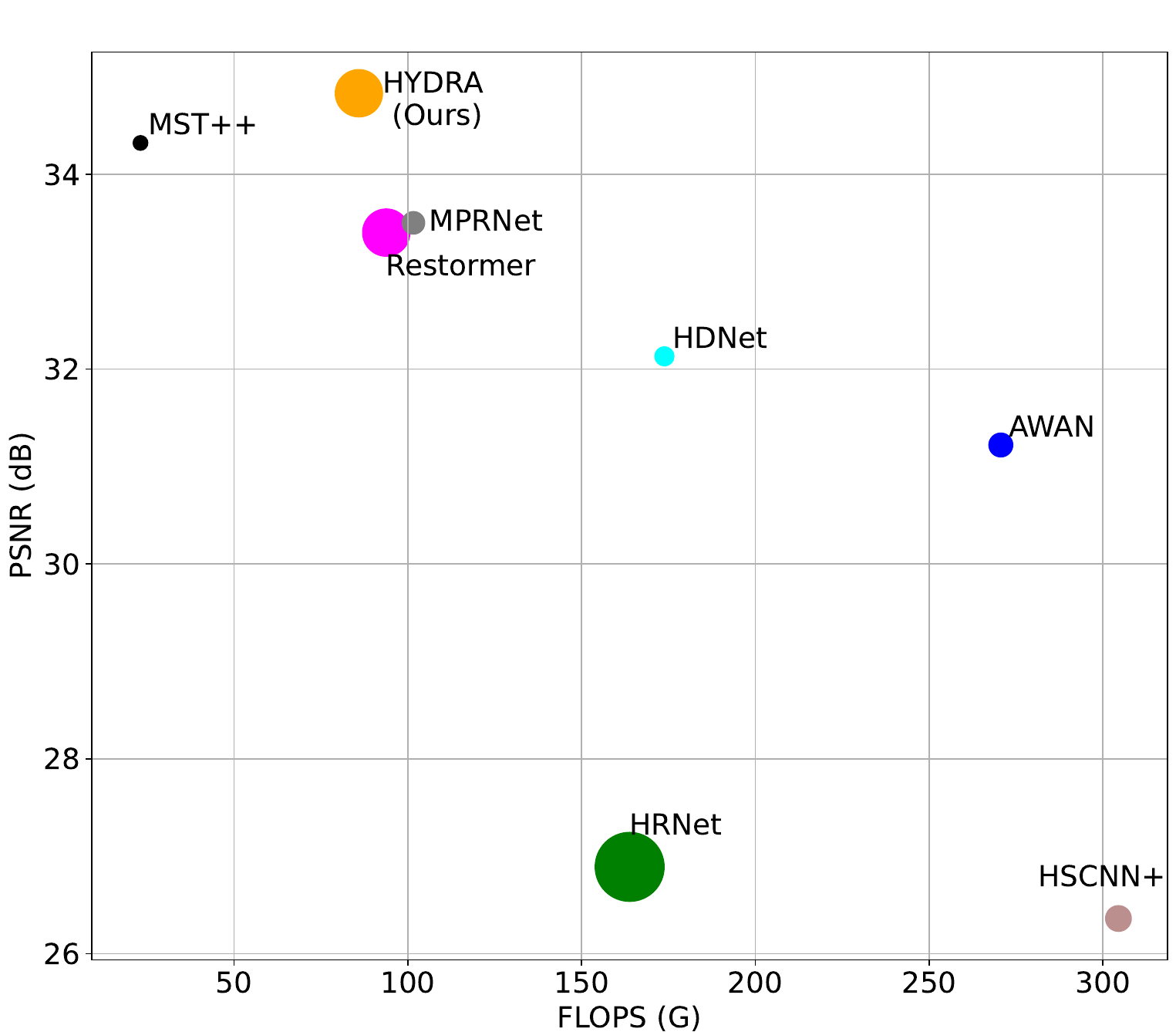}\hfill
  \includegraphics[width=0.32\textwidth]{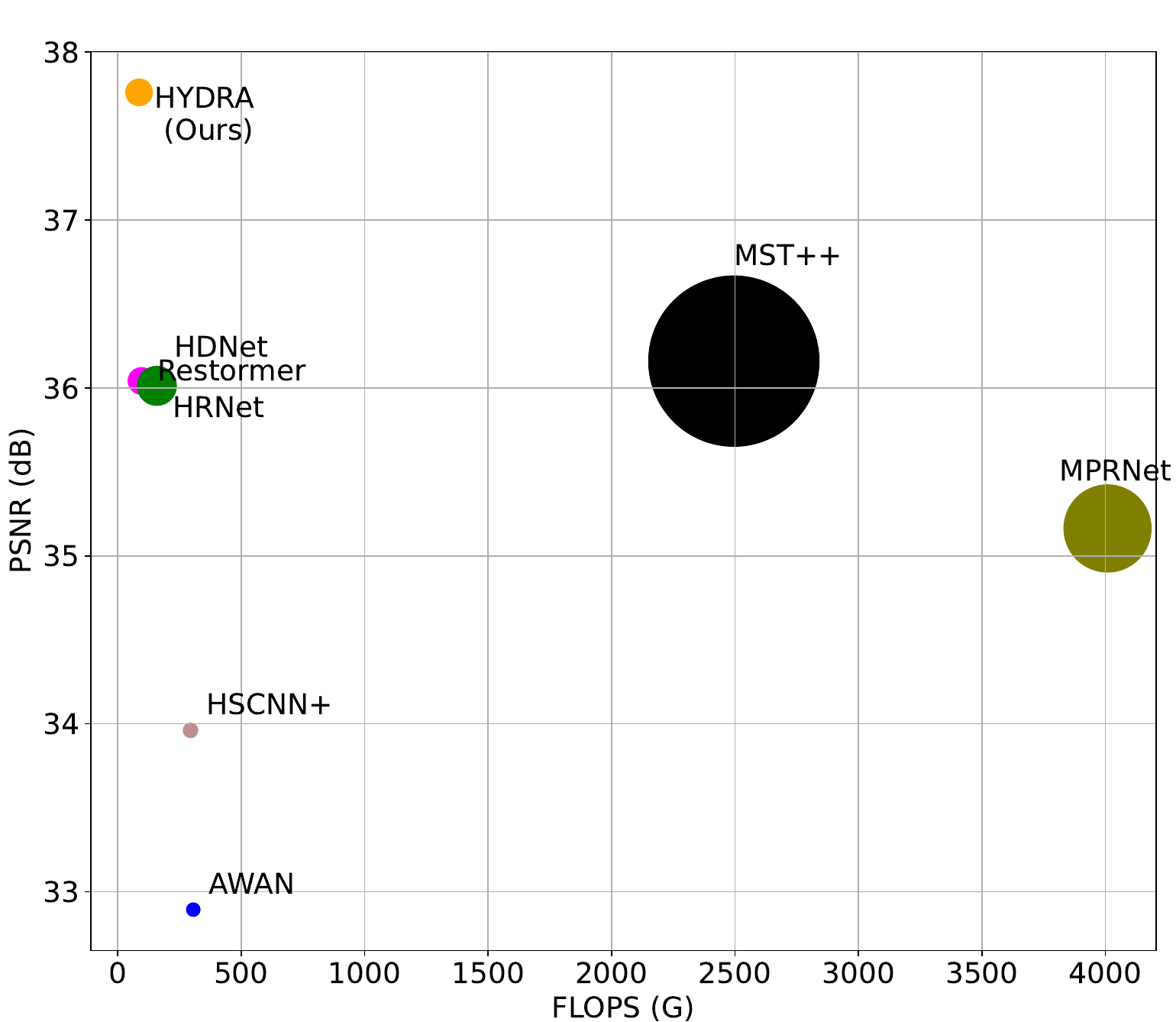}\hfill
  \includegraphics[width=0.32\textwidth]{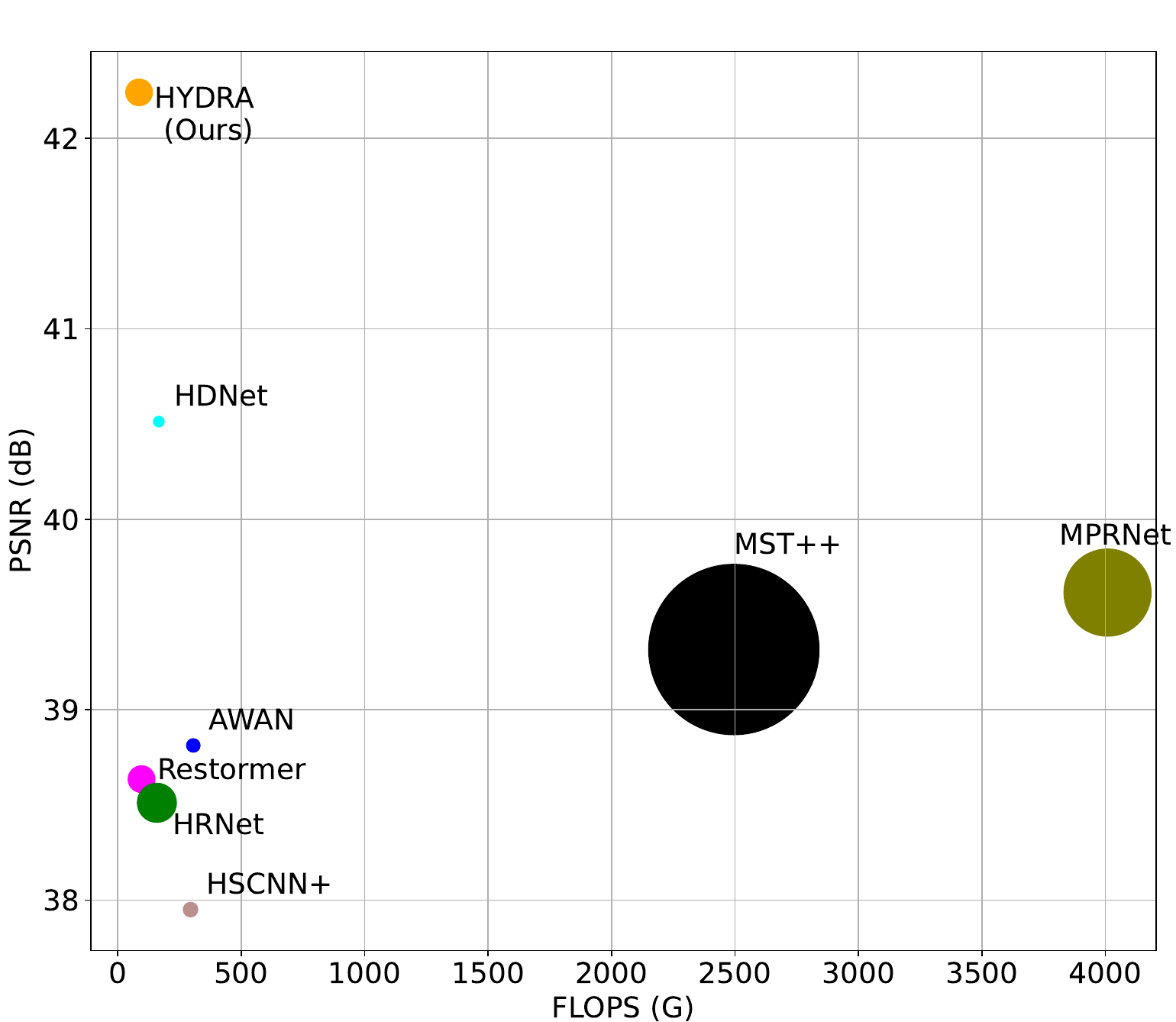}
  % \\[1ex] % small vertical gap
  \vspace{-2mm}
  \captionof{figure}{%
    (a) NTIRE-2022\quad
    (b) HySpecNet-11k\quad
    (c) LIB-HSI. HYDRA excels in spectral reconstruction across various channel depths, 
    balancing computational efficiency (FLOPS) with high performance (PSNR). 
    Circle size indicates memory cost.
  }
  \label{fig:teaser}
  \vspace{-2mm}
\end{strip}

\input{sec/0_abstract}    
\input{sec/1_intro}
\input{sec/2_litrev}
\input{sec/3_method}
\input{sec/4_experiments}
% \input{sec/4_5_limitations}
\input{sec/5_conclusion}
{
    \small
    \bibliographystyle{ieeenat_fullname}
    \bibliography{main}
}

\end{document}

%% file: sec/0_abstract.tex
\begin{abstract}
% \vspace{-2mm}
Hyperspectral images (HSI) promise to support a range of new applications in computer vision.
Recent research has explored the feasibility of generalizable Spectral Reconstruction (SR), the problem of recovering a HSI from a natural three-channel color image in unseen scenarios.
  However, previous Multi-Scale Attention (MSA) works have only demonstrated sufficient generalizable results for very sparse spectra, while modern HSI sensors contain hundreds of channels.
  This paper introduces a novel approach to spectral reconstruction via our HYbrid knowledge Distillation and spectral Reconstruction Architecture (HYDRA).
  Using a Teacher model that encapsulates latent hyperspectral image data and a Student model that learns mappings from natural images to the Teacher's encoded domain, alongside a novel training method, we achieve high-quality spectral reconstruction. 
  This addresses key limitations of prior SR models, providing SOTA performance across all metrics, including an 18\% boost in accuracy, and faster inference times than current SOTA models at various channel depths.
  % HYDRA achieves a PSNR improvement of  dB over 
  % \vspace{-1mm}
\end{abstract}

%% file: sec/1_intro.tex
\section{Introduction}
\label{sec:intro}
\vspace{-1mm}
Hyperspectral Imaging (HSI) extends traditional Red, Green, and Blue (RGB) imaging by capturing a broader spectrum across numerous narrow bands, offering detailed wavelength information invaluable in fields like medical imaging\cite{karim2023hyperspectral} and remote sensing\cite{ravikanth2017hsiextraction}.
% This capability is invaluable in fields like medical imaging and remote sensing. 
% Despite its potential, conventional HSI acquisition methods, based on spectrometers, are often impractical for dynamic or real-time applications due to their time-consuming nature and the high costs of HSI cameras.
However, conventional HSI acquisition, reliant on spectrometers, is often impractical for dynamic or real-time applications due to its time-intensive nature and the high costs of HSI cameras.
Emerging technologies in Snapshot Compressive Imaging (SCI) and computational reconstruction algorithms have started to address these challenges \cite{xing_SCI, hao_sci}. 
However, HSI still demands costly hardware and long exposures.
A promising alternative is spectral reconstruction (SR) from RGB images: it leverages standard cameras to estimate hyperspectral data, slashing costs and acquisition time.
SR can’t replace a true HSI camera everywhere, yet it is already an industrial tool for multispectral systems that recover far fewer bands.
However, SR often serves as a bridge, allowing models trained on one sensor to be deployed on another or facilitating rapid filtering during active exploration tasks.
Traditional SR methods are based on sparse coding; as such, they have limited representational ability for this task and have been mostly tested on single scenes split into multiple images, such as Salinas or Indiana Pines \cite{PURR1947}.
Advances in deep learning, especially deep Convolutional Neural Networks (CNNs), have provided initial attempts at a generalizable mapping from RGB to HSI data. 
However, these CNN-based methods struggle with capturing long-range dependencies and inter-spectral similarities.
When applying current SOTA approaches directly to SR tasks there are two primary challenges: first, an inefficiency in capturing spatial interdependencies in HSIs; second, the computational intensity of global Multi-head Self Attention (MSA), which scales quadratically, and the limitations of local window-based MSA. 
This is demonstrated in Fig.\ref{fig:teaser} where increasing the channel depth of previous techniques leads variously to dramatic increases in model size, FLOPS and inference times or a decrease in accuracy. 
These remarks are echoed by the parameter count and computational complexity increasing logarithmically for larger channel datasets that SOTA models have not been tested on.

Our work introduces HYDRA, to solve these flaws while improving the overall performance.
This is done by reformulating the SR problem to operate in a learned latent space with drastically reduced dimensionality. 
To this end, HYDRA employs a knowledge distillation framework to perform SR from RGB images via the latent HSI space of a Teacher model.
The HYDRA architecture is based on a cross-modal Teacher-student architecture, which is yet to be explored in the field of SR. 
This approach diverges from standard knowledge distillation methods by using different input data modalities in a three-step process: 
First, the teacher model is trained as a compressive autoencoder.
Second, a student spectral reconstruction model is trained to convert RGB images to latent codes matching those of the teacher model.
Finally, a further training step in both models we call the `refinement' step. 

The HSI input modality is reserved solely for the Teacher model training phase, where it is learnt channel-wise to exploit spectral information in a latent space via unsupervised training.
Our Student model instead operates on RGB and uses spatial information alongside channel-wise attention mechanisms.
As a result, HYDRA demonstrates a remarkable capability to efficiently handle generalisable SR of large datasets containing a significantly higher number of channels compared to current SOTA models, which have only been tested on channel depths of around 31.
This novel approach outperforms existing methods at a fraction of the complexity and model sizes as seen in Fig. \ref{fig:teaser}.
The main contributions of our work are as follows:
% \vspace{-1mm}
\begin{enumerate}
% \vspace{-2mm}
 \item Introduction of HYDRA, a novel approach enhancing generalizable spectral reconstruction efficiency with a unique cross-modal knowledge distillation. 
% \vspace{-2mm}
 \item An SR architecture combining modern attention mechanisms with squeeze-excitation blocks for improved computational efficiency at high channel depth.
% \vspace{-2mm}
\item The first SR benchmark and evaluation protocol for the spectral reconstruction of high dimensional HSI datasets, to support further developments in the field.
\end{enumerate}

%% file: sec/2_litrev.tex
\section{Related Work}
\label{sec:related}
Generalizable spectral reconstruction (SR) from RGB to hyperspectral images (HSIs) is challenging due to the wide spectral range of HSIs contrasted with the narrower RGB spectrum \cite{arad2016sparse, arad2020ntire, arad2022ntire}. Earlier methods used specialized RGB cameras and machine learning techniques to address the complex RGB-HSI relationship \cite{goel2015hypercam, agahian2008reconstruction, arad2016sparse}.

\subsection{Deep Learning for SR}
\label{subsec:DLSR}
Supervised SR models like those by Nguyen et al. \cite{nguyen2014training} and Robles-Kelly \cite{robles2015single} introduced white-balancing normalization and sparse coding. However, these model-based methods have limited representation capacities and poor generalization. Deep learning, particularly using convolutional neural networks (CNNs) and generative adversarial networks (GANs), has revolutionized hyperspectral reconstruction. Vision transformers have been adapted for hyperspectral SR, improving long-range dependency handling \cite{arad2022ntire, hu2022fusformer}. Models like HSCNN+ \cite{xiong2017hscnn} and MST++ \cite{cai2022mst} enhance accuracy and incorporate spectral attention. However, global transformers' computational demands scale quadratically with spatial and channel dimensions
(see for example MST++ in Fig. \ref{fig:teaser} and Tab. \ref{tb:hyspecnet11k} and Tab. \ref{tb:libhsi}), While local transformers have limited receptive fields, which hinders effective self-attention \cite{arad2022ntire, cai2022mst}. Despite advancements, these methods still require extensive training data and accurate labelling. Previous approaches are bottlenecked in computational capacity and generalization, especially at higher channel depths (see Tables \ref{tb:libhsi} and \ref{tb:hyspecnet11k}). Datasets like NTIRE 2022 \cite{arad2022ntire} and ICVL \cite{arad2016sparse} offer complex data but remain shallow in channel depth, which is impractical in modern hyperspectral applications. Many SR models are untested on deeper datasets recently released \cite{habili2022lib, fuchs2023hyspecnet}, which offer broader spectral coverage for diverse scenes. We aim to bridge this gap by enhancing SR performance across diverse datasets, including those with up to 204 channels like LIB-HSI \cite{habili2022lib} and HySpecNet-11k \cite{fuchs2023hyspecnet}.

\subsection{HSI Compression}
\label{subsec:compression}
% Hyperspectral imagery, prone to significant spatial and spectral redundancy, offers ample compression opportunities. 
% Lossless methods like quantization and entropy coding provide accurate signal reconstruction but limited size reduction \cite{mielikainen2012lossless, matnoor2013investigation}, while modern lossy techniques offer greater reductions with reasonable quality.
% HSI compression uses inter-band and intra-band methods to decrease spectral \cite{chang2013hyperspectral} and spatial redundancies \cite{zhao2016lossy}, respectively. Combined, they improve performance but may extend inference times. 
% Principal Component Analysis (PCA) minimizes spectral correlation, and used by JPEG2000 standards for HSI compression \cite{blanes2010cost, du2007hyperspectral, wang2009lossy, penna2007transform}.
% The introduction of SEBlocks blocks significantly advances feature extraction in lossy compression, optimizing neural network efficacy. 
% Tensor decomposition aids in reducing dimensions while preserving spatial details \cite{zhang2015compression}. Techniques like the Discrete Cosine Transform (DCT) and 3D-DCT address spatial and spectral aspects but may cause artifacts mitigated by wavelet transforms \cite{qiao2014effective, rasti2012hyperspectral}. 
% These methods have boosted image quality and compression ratios in applications like FuSENet \cite{roy2020fuseNet}. Our approach combines these innovations to transform spectral signatures into a lower-dimensional space, aiming to surpass existing lossy methods in efficiency.
Hyperspectral imagery's spatial and spectral redundancies offer ample compression opportunities. 
While lossless methods provide accurate reconstruction with limited size reduction \cite{mielikainen2012lossless}, lossy techniques offer greater reductions with acceptable quality. 
Inter-band and intra-band methods reduce spectral \cite{chang2013hyperspectral} and spatial \cite{zhao2016lossy} redundancies, respectively, improving performance but potentially increasing inference time. 
JPEG2000 HSI compression standards \cite{blanes2010cost, du2007hyperspectral} exploit PCA to minimize spectral correlation. 
Unlike PCA-based compression, which assumes linear relationships, HYDRA’s Teacher model uses a non-linear latent space to better encapsulate hyperspectral variations.
Tensor decomposition reduces dimensions while preserving spatial details \cite{zhang2015compression}. 
Techniques like DCT and 3D-DCT address spatial and spectral aspects but may introduce artefacts, mitigated by wavelet transforms \cite{qiao2014effective}. 
These methods have improved image quality and compression ratios in applications like FuSENet \cite{roy2020fuseNet}. 
Our approach combines these innovations to transform spectral signatures into a lower-dimensional space, aiming to surpass existing lossy methods in efficiency.

\subsection{Teacher-Student Architectures}
\label{subsec:KD}
Knowledge Distillation (KD) \cite{hinton2015distilling} efficiently trains lightweight deep learning models. In KD, a complex `teacher' model transfers knowledge to a simpler `student' model, enhancing efficiency. Traditional KD methods often require extensive datasets and computational resources \cite{ba2014deep}, leading to generalization issues. Recent advancements expand Teacher-Student architectures to knowledge expansion, adaptation, and multi-task learning \cite{gou2021knowledge, wang2021knowledge}. However, balancing teacher and student complexities remains challenging \cite{kang2021exploring}, prompting research into more efficient designs. In computer vision, these architectures enable compact models to replicate larger models' performance, crucial for resource-limited deployments. Adaptability issues persist in diverse or data-scarce environments \cite{romero2015fitnets}. Our HYDRA architecture introduces a transformative training methodology, surpassing traditional single-modality techniques. The student and teacher operate on different modalities with varying supervision and data availability. This is particularly important in precision-critical fields like hyperspectral image processing \cite{li2016learning}.
Most Teacher-Student frameworks operate within the same data modality, whereas HYDRA uniquely leverages cross-modal learning to bridge RGB and hyperspectral data.

%% file: sec/3_method.tex
\section{Method}
\label{sec:Method}
Fig.~\ref{fi:ModelOverview} illustrates the HYDRA model's architecture. The Teacher model compresses and reconstructs HSI spectra without compromising spectral detail, effectively encapsulating the hyperspectral camera's sensitivity function within its latent space. The Student model maps RGB inputs to this latent space, ensuring efficient optimisation over long-range channel image modalities while bounding the inference prediction space for more accurate spectral reconstruction. 
By optimizing the Student model within this feature-dense and regularized space, the loss function becomes more meaningful during training, facilitating the learning process and reducing the likelihood of significant errors due to predictions outside the latent space.
Noisy spectral regions are unpredictable and challenging for existing models to regress from full HSI, often resulting in smoother representations in these areas (seen in Fig. \ref{fig:pixelRecon}). 
By leveraging the regularized latent space that smooths out these noisy bands, the Student network can learn more effectively during training. Moreover, during inference on unseen data, the embedded sensitivity function aids in predicting the overall spectral shape, even if specific band intensities are inaccurate.
We detail the Teacher model in Section~\ref{subsec:Teacher}, the Student model in Section~\ref{subsec:Student}, and the HYDRA training procedure in Section~\ref{subsec:training}.
    
\subsection{Teacher model}
\label{subsec:Teacher}
\begin{figure}[t]
	\centering
	\includegraphics[width=\columnwidth]{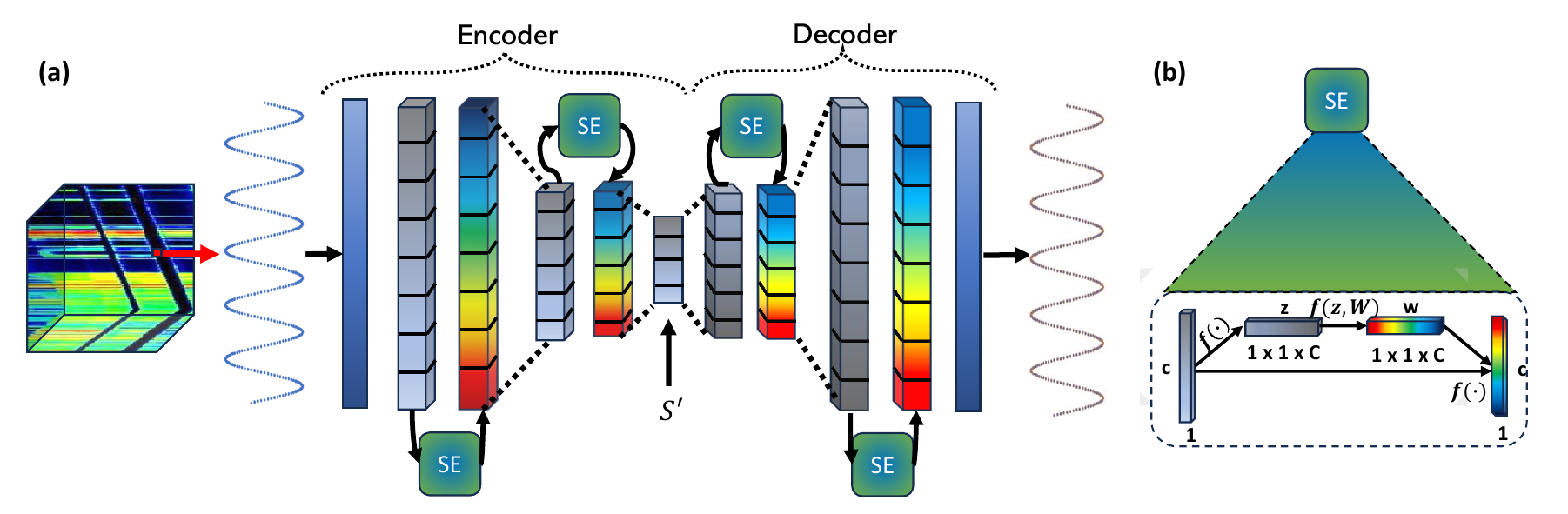}
	\caption{(a) The Teacher model, (b) the operation of an SE block.}
	\label{fi:Compression_model}
 \vspace{-7.5mm}
\end{figure}

The Teacher model of HYDRA is an unsupervised autoencoder defined generally as an encoder and decoder such that $S \approx \text{Dec}(\text{Enc}(S))$. It employs squeeze-excitation (SEBlocks) \cite{hu2017squeezeexblocks} for high-quality, pixel-wise compression.
The choice of SEBlocks over other attention mechanisms, such as Transformers or Convolutional Block Attention Modules, is motivated by their proven effectiveness in enhancing channel interdependencies with increasing channel depth without significantly increasing computational complexity.
% SE blocks were selected due to their ability to efficiently enhance channel dependencies without introducing significant computational overhead, which is critical for high-channel-depth hyperspectral data
The encoder is comprised of 1D convolution paired with 1D max-pooling layers and squeeze-excitation modules, establishing a high compression ratio (CR) for different channel depths. 
This ensures high-fidelity decoding by the mirrored decoder, thus maintaining the integrity of the compressed spectra.
% Since the model is trained on pixel wise data 
The detailed architecture of the Teacher network is shown in Fig. \ref{fi:Compression_model}.

It is worth noting that these operations function in 1D across the spectral (i.e. channels) dimension. 
While spatial information could also be captured in the Teacher model, we found this contextual information was harder to generalise given the small dataset sizes currently available compared to non-HSI task datasets. Furthermore, in development, we found it makes edges and structures of output images blurred around low-frequency details.\looseness=-1

\subsubsection{Encoder Structure}
The encoder converts a 1-dimensional input signal $S$ taken from a single pixel of an HSI containing b samples into a compressed latent representation $S'$. We introduce intermediate variables to simplify the equations.

Let $F^{l} = \text{Conv}(S^{l})$ represent the output of the convolutional layer at layer $l$, and $A^{l} = E(F^{l})$ denote the attention weights computed via the Squeeze-Excitation (SE) block. 
The encoder updates are then defined as:
\begin{equation}
S^{l+1} = \text{Enc}(S^{l}) = \text{Pool}\left( F^{l} \odot A^{l} \right)
\end{equation}
Here, $\odot$ denotes element-wise multiplication, and $S^{l}$ is the output of the $l$-th layer, with $S^{N}$ being the final compressed form $S'$. 
In this context, $\text{Conv}$ denotes a 1D convolution operation, and $\text{Pool}$ is a 1D max-pooling operation that reduces the spectral dimensionality.

\subsubsection{Decoder Structure}
The decoder reconstructs the original signal $S$ from its compressed form $S'$. Similar to the encoder, we define intermediate variables for the decoder.
Let $\tilde{F}^{l} = \text{Conv}(\tilde{S}^{l})$ be the convolutional output at layer $l$, and $\tilde{A}^{l} = E(\tilde{F}^{l})$ represent the attention weights. The decoder updates are given by:
\begin{equation}
\tilde{S}^{l-1} = \text{Dec}(\tilde{S}^{l}) = \text{UpS}\left( \tilde{F}^{l} \odot \tilde{A}^{l} \right)
\end{equation}
Since these functions operate in 1D across the spectrum, the upsampling process $\text{UpS}$ increases the spectral dimensionality of the feature map using nearest-neighbour interpolation, facilitating reconstruction. The final output of the decoder is the reconstructed signal, matching the input dimensions of the encoder.
In this modular approach, the encoder and decoder share a symmetrical structure, where $\tilde{W}$, $\tilde{S}$, and associated parameters act as the decoder's counterparts to the encoder's weights and inputs. Differently from a traditional U-Net \cite{ronneberger2015unet}, this network omits skip connections between the encoder and decoder blocks to allow the decoder to operate independently of the Student network during inference.
We discovered that pixel-wise compression more effectively encapsulates hyperspectral data, subsequently refining the Student network's spectral predictions. 
% This method also simplifies representing the spectral sensitivity of the cameras used in the datasets, thereby minimizing erroneous outputs. 
These improvements are demonstrated in the heatmaps presented in Fig.~\ref{fig:heatmapHyspec}.

\subsection{Student Model}
\label{subsec:Student}
\begin{figure}[t]
    \centering
    \includegraphics[width=1.0\columnwidth]{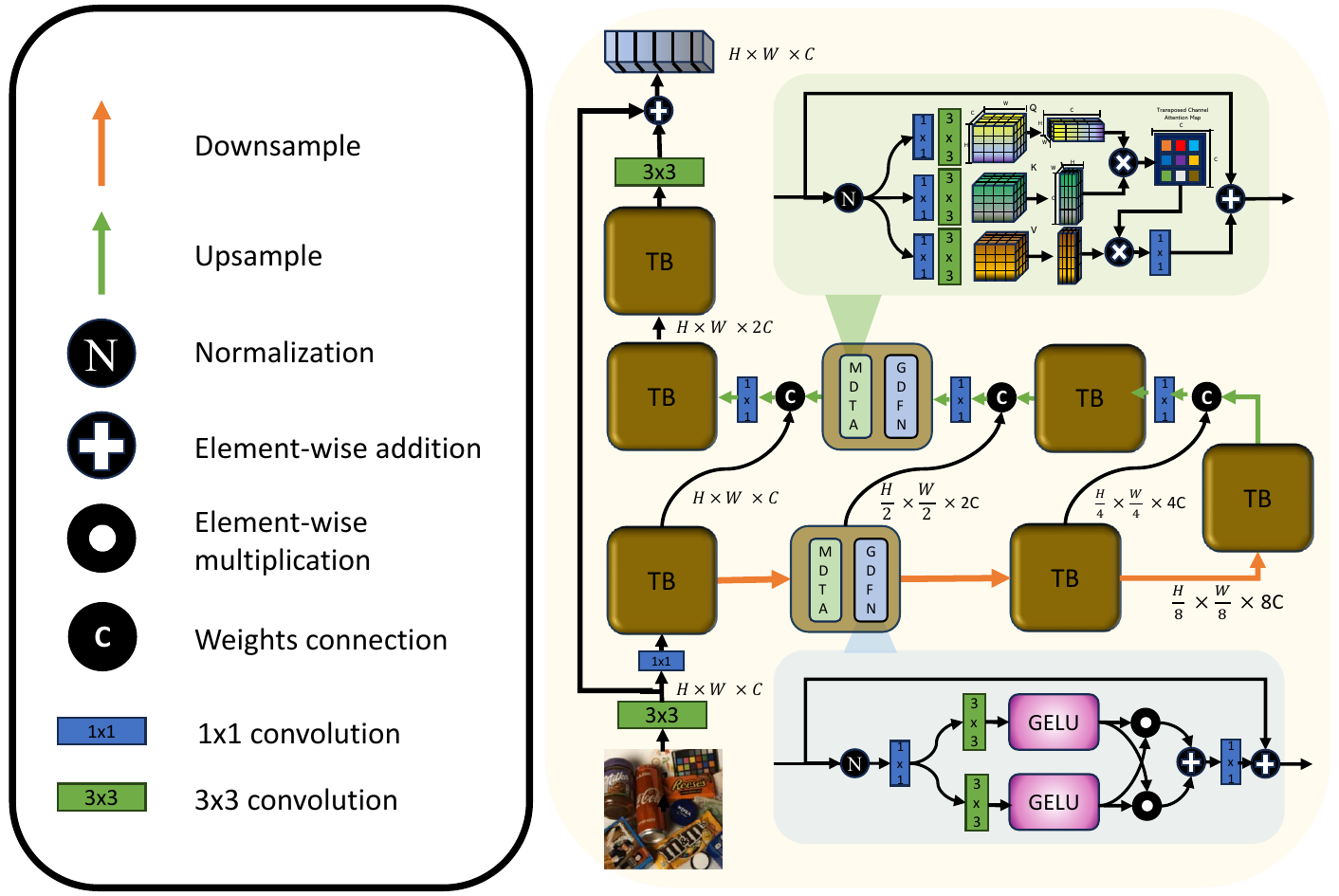}
    \caption{Our Student model: A U-Net shaped Vision Transformer network in the channel domain, utilizing Multi-Dconv Head Transposed Attention (MDTA) and Dual-Gated-Dconv Feed-Forward Network (DGDF) modules. Green 3x3 blocks are depth-wise convolutions and blue blocks 1x1 are pixel-wise convolutions to capture spatial and spectral shapes.
    }
    % \caption{Our Student Network Architecture}
    \label{fi:Transformer}
    \vspace{-2mm}
\end{figure}

Our proposed composite framework HYDRA integrates a Teacher network for efficient spectral compression and a Student network for high-fidelity spectral reconstruction. 
% The Student model leverages the latent space knowledge encapsulated in the Teacher network to aid spectral reconstruction. 
The architecture of our Student model, depicted in Fig.~\ref{fi:Transformer}, is inspired by U-Net style attention networks \cite{He2023UNetmerUM}, renowned for their channel-focused attention mechanisms, but tailored for natural image restoration tasks.

Our Student model employs the Multi-Dconv Head Transposed Attention (MDTA) \cite{zamir2022restormer} as its transformer backbone, integrating spatial convolution within the channel-wise attention mechanism for enhanced feature extraction and representation. Each block in the U-Net pipeline leverages these Transformer blocks to capture both spatial and spectral dependencies.

In the encoder, the feature extraction function \(Y(X)\) represents the process where input feature maps \(X\) are transformed by applying a series of MDTA blocks, reducing the spatial dimensions while increasing the number of channels. In the decoder, the function \(\tilde{Y}(\tilde{X})\) represents the upsampling process, where the decoder reconstructs the feature maps to their original spatial dimensions by reversing the operations of the encoder. Skip connections between corresponding encoder and decoder layers ensure that spatial details are preserved, providing rich feature representations at multiple scales. This architecture allows for efficient long-range dependency modelling, similar to the Restormer architecture. The model operation is formalised explicitly in the following subsections.

\subsubsection{U-Net Pipeline}
\label{subsubsec:stuUnet}
The Student model's U-Net-like encoder-decoder architecture processes an input image \( I \in \mathbb{R}^{H \times W \times 3} \) through four levels. The initial convolutional layer transforms the input into feature embeddings \( X^0 \in \mathbb{R}^{H \times W \times C} \), where \( C \) represents the number of channels.

In the encoder, each layer generates feature maps \( Y(X^{l-1}) = X^l \in \mathbb{R}^{H/2^l \times W/2^l \times 2^lC} \) by employing an increasing number of Transformer blocks, which utilize MDTA. This process reduces the spatial dimensions (\( H \) and \( W \)) while increasing the channel count. The decoder, using upsampling operations, reconstructs the original image dimensions, with the help of skip connections between the corresponding Student encoder and decoder layers $\tilde{Y}(\tilde{X}^l) + X^l = \tilde{X}^{l-1}$. This helps retain important features from earlier stages during the reconstruction process.

In the student decoder stage, deep features are refined through a convolutional layer, leading to the generation of the residual encoded hyperspectral image. The system outputs the final encoded image, which is a combination of the initial feature embeddings and the residual image:
\begin{equation}
    \tilde{S} = X^0 + \text{Conv}(\tilde{X}^0)
\end{equation}
Here, \(\tilde{S}\) represents the final latent restored image, and $\tilde{X}^0$ is the output of the final decoder block
The U-Net architecture \cite{ronneberger2015unet} is particularly well-suited for tasks that require detailed spatial-channel correlation, such as super-resolution or spectral-reconstruction. Its encoder-decoder structure efficiently captures both high-level and low-level features.

% The Student model's transformer backbone is the Multi-Dconv Head Transposed Attention (MDTA), which integrates spatial convolution within the channel-wise attention mechanism for enhanced feature extraction and representation. 
% Each block in the U-Net pipeline (i.e., \(Y(X)\) and \(\tilde{Y}(\tilde{X})\)) employs these Transformer blocks, allowing for efficient long-range dependency modelling.

\subsubsection{MDTA blocks}
The MDTA block is a key part of the \(Y(X)\) and \(\tilde{Y}(\tilde{X})\) functions, responsible for feature extraction within the U-Net pipeline. It operates with channel depth-wise (\( H_{\text{d}} \)) and pixel-wise (\( H_{\text{p}} \)) convolutions on layer-normalized input \(X\):
% Channel depth-wise (\( H_{\text{d}} \)) and pixel-wise (\( H_{\text{p}} \)) convolutions on layer normalized input \(X\) are defined as:
\begin{align}
    H_{\text{d}}(X) &= \mathcal{C}_D(\text{LN}(X)) \\
    H_{\text{p}}(X) &= \mathcal{C}_P(\text{LN}(X))
\end{align}
where \(\text{LN}\) denotes layer normalization.

The combined operation of depth-wise and pixel-wise convolutions is then defined as:
\begin{equation}
    \mathcal{C}_{\text{dp}}(X) = H_{\text{d}}(H_{\text{p}}(X))
\vspace{-2mm}
\end{equation}

The Student model, unlike the Teacher model, emphasizes spatial context across RGB images. It employs MSA blocks for channel-wise attention calculation, coupled with feed-forward blocks to effectively propagate feature transformations through the network.

The attention mechanism operates by computing a dot-product interaction of the query (\(Q\)), key (\(K\)), and value (\(V\)) outputs to generate the attention map:
\begin{equation}
{
    A(X) = \mathcal{C}_{\text{dp}}^{V}(X) \cdot \text{Softmax}\left(\frac{\mathcal{C}_{\text{dp}}^{K}(X) \cdot {\mathcal{C}_{\text{dp}}^{Q}(X)}^{\top}}{\alpha}\right),
 }
    \end{equation}
where \( \alpha \) is a learnable scaling parameter.
The enhanced feature representation \( MDTA \) is generated by combining the attention map \( A \) with the input feature \( X \):
\begin{equation}
    \text{MDTA}(X) = H_{\text{p}}(A(X)) + X.
    \vspace{-1mm}
\end{equation}
The combination of depth-wise and pixel-wise convolutions in MDTA allows for a more nuanced and effective feature transformation. Depth-wise convolutions handle channel-wise interactions efficiently, while pixel-wise convolutions focus on spatial details, making the model adept at capturing both spatial and spectral features.

\subsubsection{DGFN} 
The Dual Gated-Dconv Feed-Forward Network processes the output from preceding layers, focusing on the effective propagation of texture features. The DGFN incorporates a dual gating mechanism defined as:
\begin{equation}
    M^{g}_G(X) = \phi(\mathcal{C}_{\text{dp}}(\text{MDTA}(X)))
\end{equation}
where \( g \) is either the first (1) or second (2) gate, and \( \phi \) represents an activation function.

These pathways are combined in the encoder branch as follows:
\begin{equation}
    Y_G(X) = (M^1_G(X) \odot M^2_G(X)) + (M^1_G(X) \odot M^2_G(X))
\label{eq:DGFN}
\end{equation}
\begin{equation}
{
    Y(X) = H_{\text{p}}(Y_G(X)) + \text{MDTA}(X).
}
\end{equation}
This approach merges gated features with the original features, ensuring the preservation of essential information while emphasizing critical textural elements.
This representation is then fed into the decoder of the Teacher model for SR. The addition of a skip connection from the encoder enhances the Student decoder's ability to recover fine details lost during compression. The Student decoder operation, along with the skip connection from the encoder, is formalized as:
\begin{equation}
{
    \tilde{Y}(\tilde{X}) = H_p(\tilde{Y}_G(\tilde{X})) + \text{MDTA}(\tilde{X}) + X
}
\end{equation}
where \( \tilde{X} \) represents the input to the Teacher's decoder, \( H_p \) denotes pixel-wise convolutions, \( \tilde{Y}_G \) is the Dual Gated-Dconv Feed-Forward Network as in eq.\ref{eq:DGFN}, and \( X \) is the feature map from the equivalent layer of the Student's encoder provided via the skip connection. This output is then passed to the Teacher's decoder, which reconstructs the final hyperspectral image. To clarify, the Student processes RGB to latent space, and the Teacher converts this latent space into hyperspectral data.
% This collaborative approach between the Student and Teacher models ensures the effective translation of RGB images into accurate hyperspectral reconstructions.

\subsection{Training Procedure} 
\label{subsec:training}
\begin{figure}[t]
	\centering
	\includegraphics[width=\columnwidth]{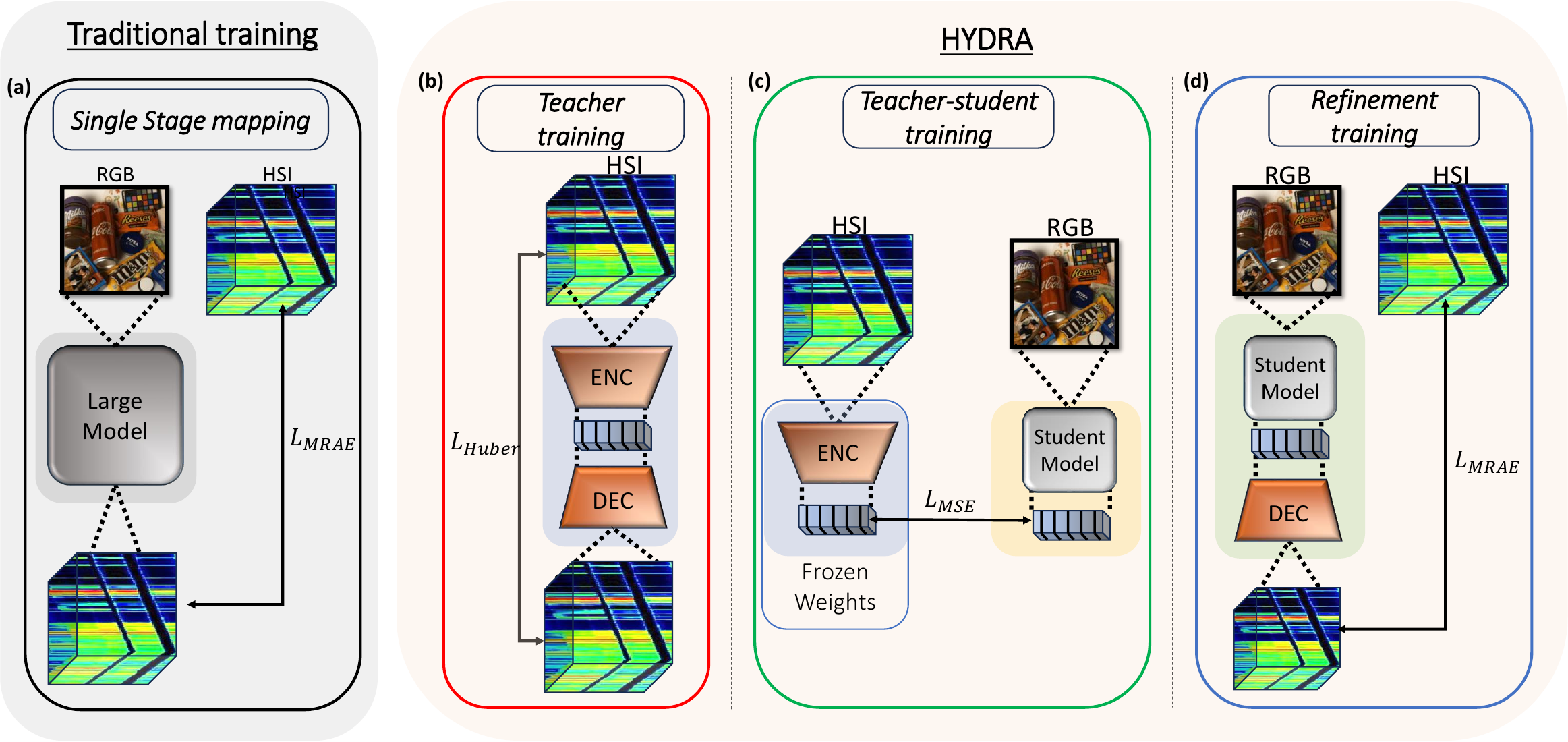}
     \vspace{-7mm}
	\caption{(a) The traditional training method for SR. (b), (c) and (d) show the stages of the HYDRA training procedure. }
	\label{fi:ModelOverview}
 \vspace{-3mm}
\end{figure}

The training procedure outlined in this section is illustrated via Fig.~\ref{fi:ModelOverview}. 
Our novel HYDRA training methodology uses a three-stage process.

\paragraph{Training stage one:}
The first stage trains the Teacher network (section \ref{subsec:Teacher} and Fig. \ref{fi:ModelOverview}.a) to compress HSIs into a compact representation, providing a feature dense latent space for the Student network. 
The loss function used is the Huber loss which provides a smooth loss throughout training, as it handles outliers effectively while maintaining smooth gradients. 

\paragraph{Training stage Two:}\vspace{-2mm}
In the second stage, the Student network learns to map RGB images to the frozen Teacher's latent HSI representations (see Fig. \ref{fi:ModelOverview}(c)). The loss function, mean absolute error (MAE), is defined as:
\begin{equation}
{
    L_{sr} = \frac{1}{n} \sum_{i=1}^{n} |\tilde{Y}_i - S'_i|
    \label{eq:MAE}
}
\vspace{-2mm}
\end{equation}
where $\tilde{Y}_n$ represents the Student's predictions and $S'$ is the Teachers latent representation and $n$ is number of data sample. We choose MAE because it penalizes all errors equally, making it robust to outliers. Unlike squared loss, which disproportionately emphasizes larger errors, MAE treats all deviations uniformly. This helps the Student network minimize discrepancies with the Teacher's outputs across latent spectral channels without being overly influenced by outliers, leading to more stable training.

\paragraph{Training stage Three:} \vspace{-2mm}
The final stage involves fine-tuning both the Teacher's decoder and the Student model together (Fig. \ref{fi:ModelOverview}.c), To this end, we unfreeze the Teachers decoder, and train the system end to end using an MSE loss to minimize the difference between the predicted and ground truth HSI.
% This phase enhances model performance by allowing the latent space to adapt to be better suited to the end goal of spectral reconstruction. 
The refinement step aligns the latent space with the end goal of spectral reconstruction, addressing minor discrepancies that may arise due to independent training in earlier stages.

%% file: sec/4_experiments.tex
\vspace{-2mm}
\section{Experiments}
\label{sec:experiments}

\begin{figure*}[t]
  \centering
  \begin{subfigure}[b]{\textwidth}
    \includegraphics[width=\linewidth]{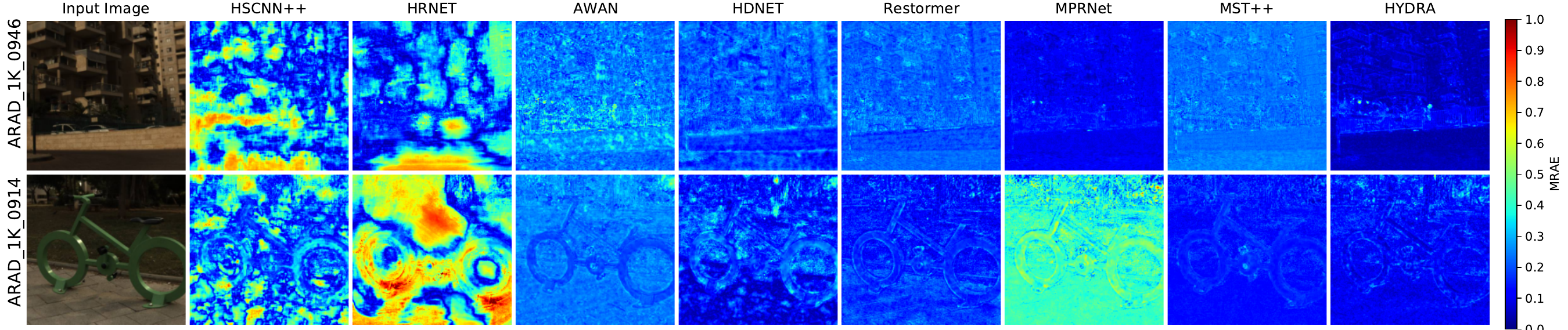}

    \caption{NTIRE-2022 dataset examples}
    \label{fig:heatmapARAD}
  \end{subfigure}

  \begin{subfigure}[b]{\textwidth}
    \includegraphics[width=\linewidth]{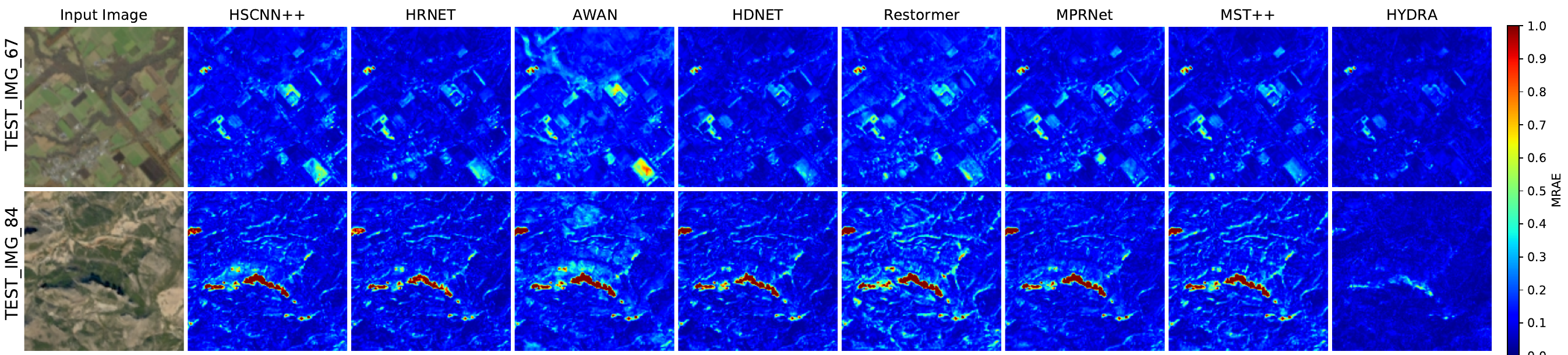}
    \caption{HySpecNet-11k dataset examples}
    \label{fig:heatmapHyspec}
  \end{subfigure}

  \begin{subfigure}[b]{\textwidth}
    \includegraphics[width=\linewidth]{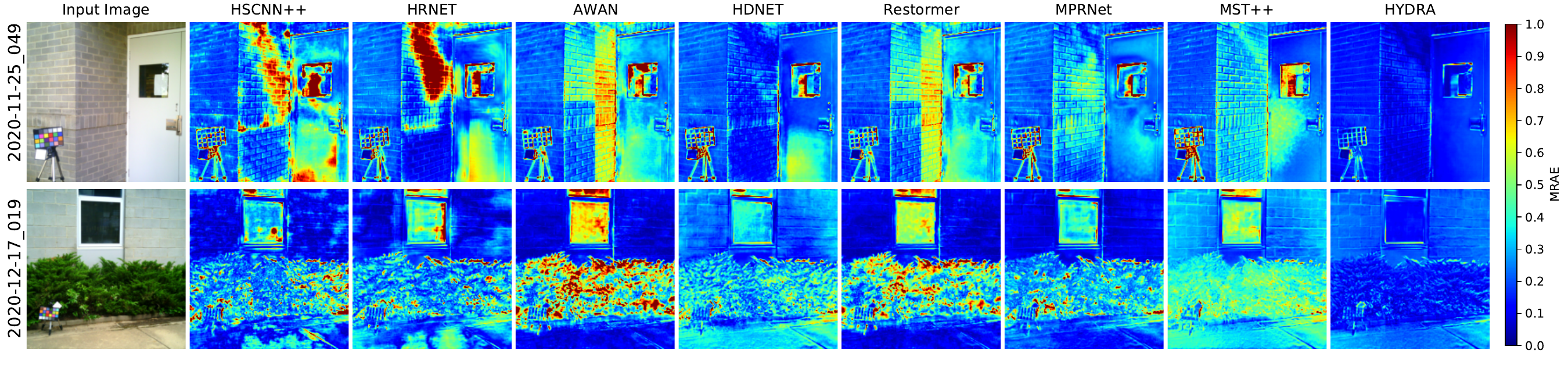}
    \caption{LIB-HSI dataset examples} 
    \label{fig:heatmapLIBHSI}
  \end{subfigure}

  % General caption for the whole figure
  \vspace{-4mm}
  \caption{MRAE heatmap errors for different datasets.}
  \label{fig:heatmaps}
  \vspace{-5mm}
\end{figure*}

We trained our models on a single NVIDIA A100 80GB GPU, though HYDRA remains compatible with smaller GPUs due to its low memory usage. We used the ARAD1K/NTIRE2022 \cite{arad2022ntire}, LIB-HSI \cite{habili2022lib}, and HySpecNet11K \cite{fuchs2023hyspecnet} datasets with their specified training splits. All methods were implemented in PyTorch, utilising the recommended optimisation and learning rates from their original papers, with minor noise removal adjustments made for the larger LIB-HSI and HySpecNet11K datasets. Each model was trained for 300 epochs with a batch size of 20, and we report the average accuracy per dataset to provide consistency.

As seen from Tables \ref{tb:ntire22}, \ref{tb:hyspecnet11k}, and \ref{tb:libhsi}, HYDRA outperforms existing current SOTA SR models in the SR task. 
This improvement is attributed to our innovative Teacher-Student framework, which efficiently leverages a compact latent space for the Student model to explore. 
This effect provides a defined space that improves optimisation via a bounded space via a more descriptive latent space modality. Additionally, this modality is inherently smoother and avoids regression over noisy areas of spectra, which can be handled more elegantly via the Teacher network's latent space. 
In our quantitative analysis (refer to Tables \ref{tb:ntire22}, \ref{tb:hyspecnet11k}, \ref{tb:libhsi} and Fig. \ref{fig:teaser}), our approach exhibits significant improvements across all three accuracy evaluation metrics and inference times against other transformer models (Table \ref{tb:transformer_times}, particularly in datasets with far deeper channel depths like HyspecNet-11k and LIB-HSI being over six times that of NTIRE-22.
Notably, we maintain a stable parameter count and FLOPS compared to other methods. Methods with smaller parameters or FLOPS tend to perform poorly across all metrics. FLOPS are computed using the fvcore library for consistency. In HYDRA, the Teacher’s decoder is included with the Student during inference to calculate total FLOPS.

HYDRA's impressive computational efficiency is due to the offloading of initial computation to a compact Teacher model, which shrinks the channel space for the MSA Student model. 
In most baseline models, larger channel depth datasets cause models to expand significantly, whereas HYDRA remains much smaller due to the reduced channel size provided by the Teacher model.
By offering a compact, lower-dimensional target, the latent space reduces the computational load during stage 2 of training, simplifying the process and facilitating more efficient learning of spectral distributions from the full architecture.
Our architecture's ability to capture complex spectral relationships through the latent space enables accurate reconstruction even in high-dimensional settings. Importantly, HYDRA maintains a stable parameter count and low FLOPS due to the Teacher's efficient spectral compression and the Student's focused reconstruction. In contrast, methods with fewer parameters or lower FLOPS cannot often model intricate spectral details, resulting in poorer performance.

HYDRA surpasses MSA-based models on complex datasets, as MSA models often struggle with high-dimensional spectral data. By effectively modelling inter-channel dependencies in the Teacher’s latent space, HYDRA guides the Student in a smoother latent space. This approach handles high-dimensional data more efficiently and robustly than other models.

These results highlight the efficacy of our Teacher-Student cross-modality approach. By mapping RGB inputs to the Teacher’s latent space, the Student is guided by rich spectral information, improving generalisation and enabling more accurate HSI reconstruction than methods without cross-modality learning.

\begin{table}[tp]
  \caption{Results on NTIRE-22 \cite{arad2022ntire} (31 channels). Best‐in‐column in \textbf{bold}, second best \underline{underlined}. The * indicates current SOTA transformer backbone methods.}
  \vspace{-2mm}
  \centering
  \resizebox{\linewidth}{!}{
    \begin{tabular}{|c|c|c|c|c|}
      \hline
      \textbf{Model} 
        & \textbf{FLOPS (G)} 
        & \textbf{MRAE} ($\downarrow$) 
        & \textbf{RMSE} ($\downarrow$) 
        & \textbf{PSNR} ($\uparrow$) 
      \\ \hline
      *MST++\cite{cai2022mst}        
        & \cellcolor{lightgreen}\textbf{23.05} 
        & \cellcolor{lightyellow}\underline{0.1645} 
        & \cellcolor{lightyellow}\underline{0.0248} 
        & \cellcolor{lightyellow}\underline{34.32} \\
      *MPRNet \cite{Zamir_2021_CVPR}     
        & 101.59         
        & 0.1817             
        & 0.0270             
        & 33.50              \\
      *Restormer  \cite{zamir2022restormer}  
        & 93.77          
        & 0.1833             
        & 0.0274             
        & 33.40             \\
      HDNet    \cite{hu2022hdnet}     
        & 173.81         
        & 0.2048             
        & 0.0317             
        & 32.13              \\
      AWAN \cite{awan}          
        & 270.61         
        & 0.2500             
        & 0.0367             
        & 31.22             \\
      HRNet   \cite{hrnet}      
        & 163.81         
        & 0.3476             
        & 0.0550             
        & 26.89             \\
      HSCNN++  \cite{shi2018hscnn}      
        & 304.45         
        & 0.3814             
        & 0.0588             
        & 26.36              \\
      \textbf{HYDRA (ours)} 
        & \cellcolor{lightyellow}\underline{85.90} 
        & \cellcolor{lightgreen}\textbf{0.1556}    
        & \cellcolor{lightgreen}\textbf{0.0221}    
        & \cellcolor{lightgreen}\textbf{34.83}    \\
      \hline
    \end{tabular}
  }
  \label{tb:ntire22}
  \vspace{-5mm}
\end{table}

\begin{table}[tp]
  \caption{Results on HySpecNet-11k \cite{fuchs2023hyspecnet} (202 channel) dataset.}
  \vspace{-2mm}
  \centering
  \resizebox{\linewidth}{!}{
    \begin{tabular}{|c|c|c|c|c|}
      \hline
      \textbf{Model} 
        & \textbf{FLOPS (G)} 
        & \textbf{MRAE} ($\downarrow$) 
        & \textbf{RMSE} ($\downarrow$) 
        & \textbf{PSNR} ($\uparrow$) 
      \\ \hline
      *MST++ \cite{cai2022mst}        
        & 2595.2 
        & 0.3133             
        & 0.0196             
        & \cellcolor{lightyellow}\underline{36.159} \\
      *MPRNet \cite{Zamir_2021_CVPR}     
        & 4007.32
        & 0.3283             
        & 0.0210             
        & 35.163             \\
      *Restormer \cite{zamir2022restormer}   
        & \cellcolor{lightyellow}\underline{96.34} 
        & 0.2960 
        & 0.0187             
        & 36.041             \\
      HDNet \cite{hu2022hdnet}         
        & 166.37 
        & 0.2856             
        & \cellcolor{lightyellow}\underline{0.0186} 
        & 36.09              \\
      AWAN \cite{awan}         
        & 306.07 
        & \cellcolor{lightyellow}\underline{0.2218} 
        & 0.0268             
        & 32.894             \\
      HRNet   \cite{hrnet}       
        & 158.57 
        & 0.2627             
        & 0.0192             
        & 36.012             \\
      HSCNN++   \cite{shi2018hscnn}    
        & 294.89 
        & 0.2953             
        & 0.0242             
        & 33.961             \\
      \textbf{HYDRA (ours)} 
        & \cellcolor{lightgreen}\textbf{86.59} 
        & \cellcolor{lightgreen}\textbf{0.1563}    
        & \cellcolor{lightgreen}\textbf{0.0160}    
        & \cellcolor{lightgreen}\textbf{37.759}    \\
      \hline
    \end{tabular}
  }
  \label{tb:hyspecnet11k}
  \vspace{-5mm}
\end{table}

\begin{table}[htp]
  \caption{Results on LIB-HSI \cite{habili2022lib} (204 channel) dataset.}
  \vspace{-2mm}
  \centering
  \resizebox{\linewidth}{!}{
    \begin{tabular}{|c|c|c|c|c|}
      \hline
      \textbf{Model}
        & \textbf{FLOPS (G)}
        & \textbf{MRAE} ($\downarrow$)
        & \textbf{RMSE} ($\downarrow$)
        & \textbf{PSNR} ($\uparrow$)
      \\ \hline
      *MST++ \cite{cai2022mst}
        & 2595.2
        & 0.4041
        & 0.01254
        & 39.316
      \\
      *MPRNet \cite{Zamir_2021_CVPR}
        & 4008.53
        & 0.3922
        & 0.01181
        & 39.133
      \\
      *Restormer \cite{zamir2022restormer}
        & \cellcolor{lightyellow}\underline{96.34}
        & 0.3905
        & 0.01309
        & 38.668
      \\
      HDNet \cite{hu2022hdnet} 
        & 166.37
        & \cellcolor{lightyellow}\underline{0.3102}
        & \cellcolor{lightyellow}\underline{0.01067}
        & \cellcolor{lightyellow}\underline{40.51}
      \\
      AWAN \cite{awan}
        & 306.07
        & 0.4230
        & 0.01248
        & 38.812
      \\
      HRNet \cite{hrnet} 
        & 158.57
        & 0.3881
        & 0.01280
        & 38.777
      \\
      HSCNN++ \cite{shi2018hscnn}
        & 294.89
        & 0.4334
        & 0.01350
        & 38.069
      \\
      \textbf{HYDRA (ours)}
        & \cellcolor{lightgreen}\textbf{86.59}
        & \cellcolor{lightgreen}\textbf{0.3004}
        & \cellcolor{lightgreen}\textbf{0.00905}
        & \cellcolor{lightgreen}\textbf{42.242}
      \\
      \hline
    \end{tabular}
  }
  \label{tb:libhsi}
  \vspace{-5mm}
\end{table}

\begin{table}[htp]
  \caption{Inference time (ms) for a single image for the three best SOTA methods and HYDRA (ours).}
  \vspace{-2mm}
  \centering
  \resizebox{\linewidth}{!}{
    \begin{tabular}{|c|c|c|c|}
      \hline
      \textbf{Model} 
        & \textbf{NTIRE-22} 
        & \textbf{HySpecNet-11k} 
        & \textbf{LIB-HSI} 
      \\ \hline
      MST++ \cite{cai2022mst}           
        & \cellcolor{lightyellow}102.52                  
        & 233.53                     
        & 3119.04               
      \\ 
      MPRNet \cite{Zamir_2021_CVPR}     
        & \cellcolor{lightgreen}74.97                  
        & 212.25                     
        & 2861.39               
      \\ 
      Restormer \cite{zamir2022restormer}
        & 147.30                  
        & \cellcolor{lightyellow}32.42                     
        & \cellcolor{lightyellow}219.98               
      \\ 
      \textbf{HYDRA (ours)}            
        & 154.11                  
        & \cellcolor{lightgreen}16.09                     
        & \cellcolor{lightgreen}172.70               
      \\ \hline
    \end{tabular}
  }
  \label{tb:transformer_times}
  \vspace{-7mm}
\end{table}

In table \ref{tb:transformer_times}, we record the time taken to do inference on a single image from each dataset for the top-performing methods in previous tests. Here, this highlights how HYDRA's lower FLOP counts translate to high-accuracy applications.

\subsection{Qualitative}
\label{sec:Qualitative}
\begin{figure*}[ht]
    \centering

    \begin{subfigure}[b]{0.33\textwidth}
        \includegraphics[width=0.95\columnwidth]{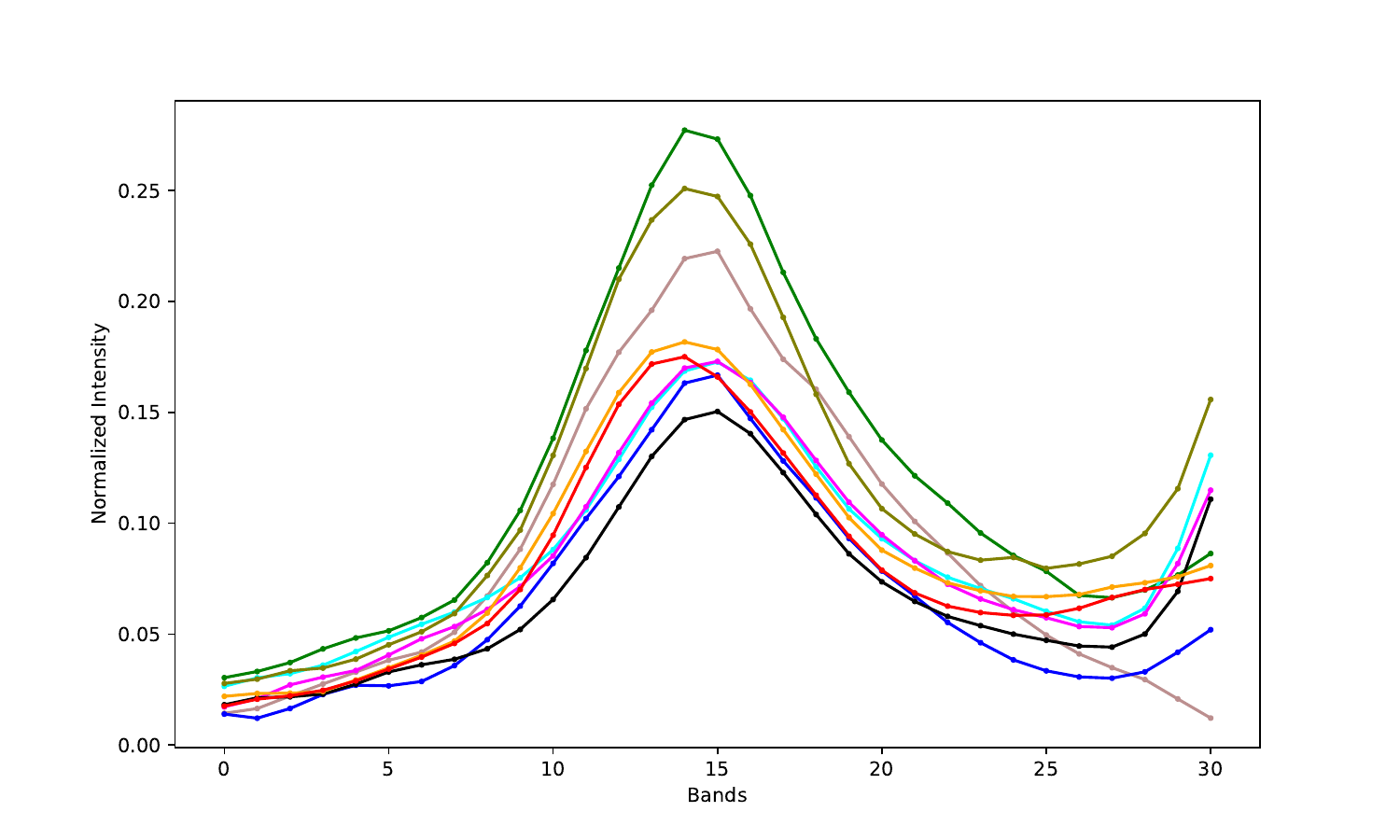}
        \caption{NTIRE-2022}
        \label{fig:arad1}
    \end{subfigure}
    \begin{subfigure}[b]{0.33\textwidth}
        \includegraphics[width=0.95\columnwidth]{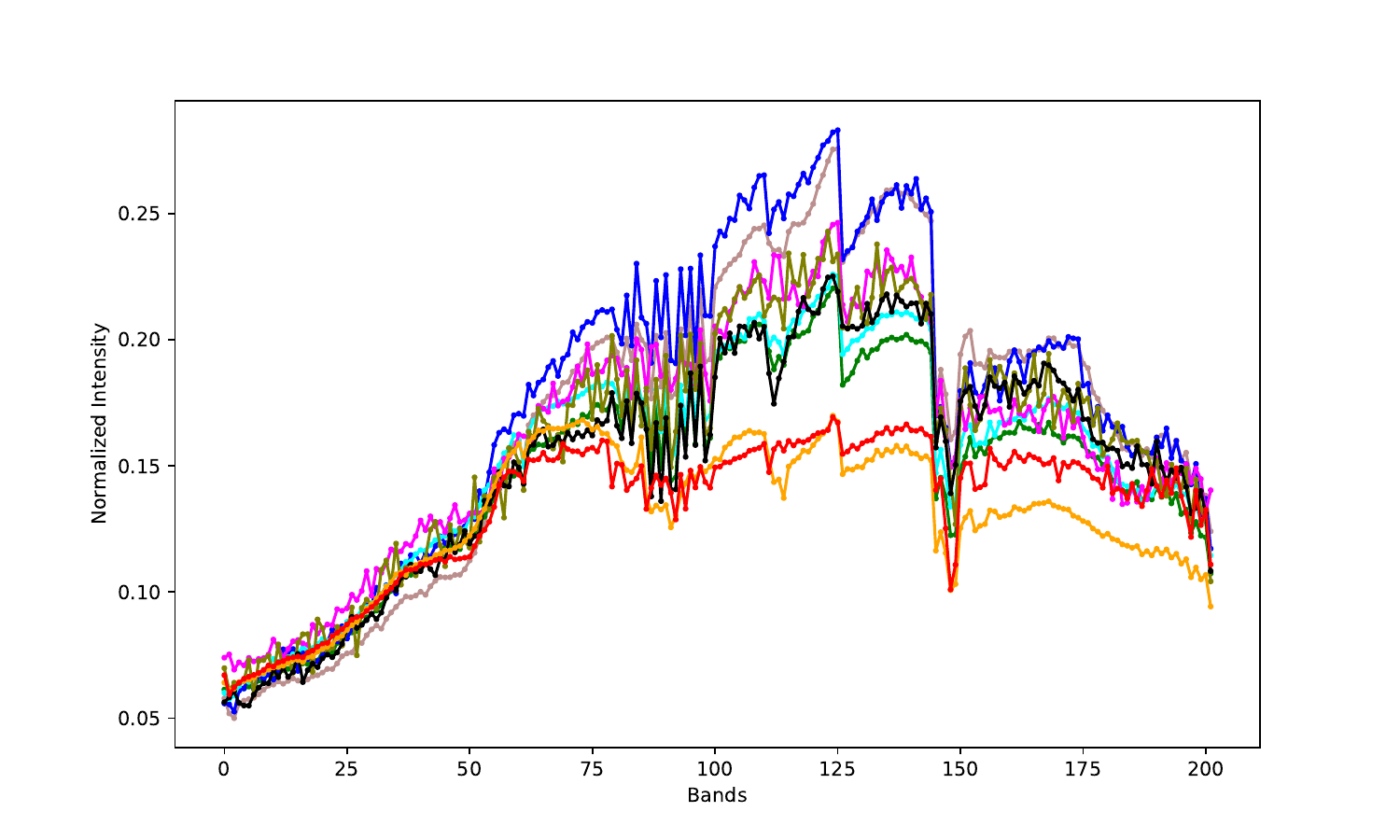}
        \caption{HySpecNet-11k}
        \label{fig:hyspec1}
    \end{subfigure}
    \begin{subfigure}[b]{0.33\textwidth}
        \includegraphics[width=0.95\columnwidth]{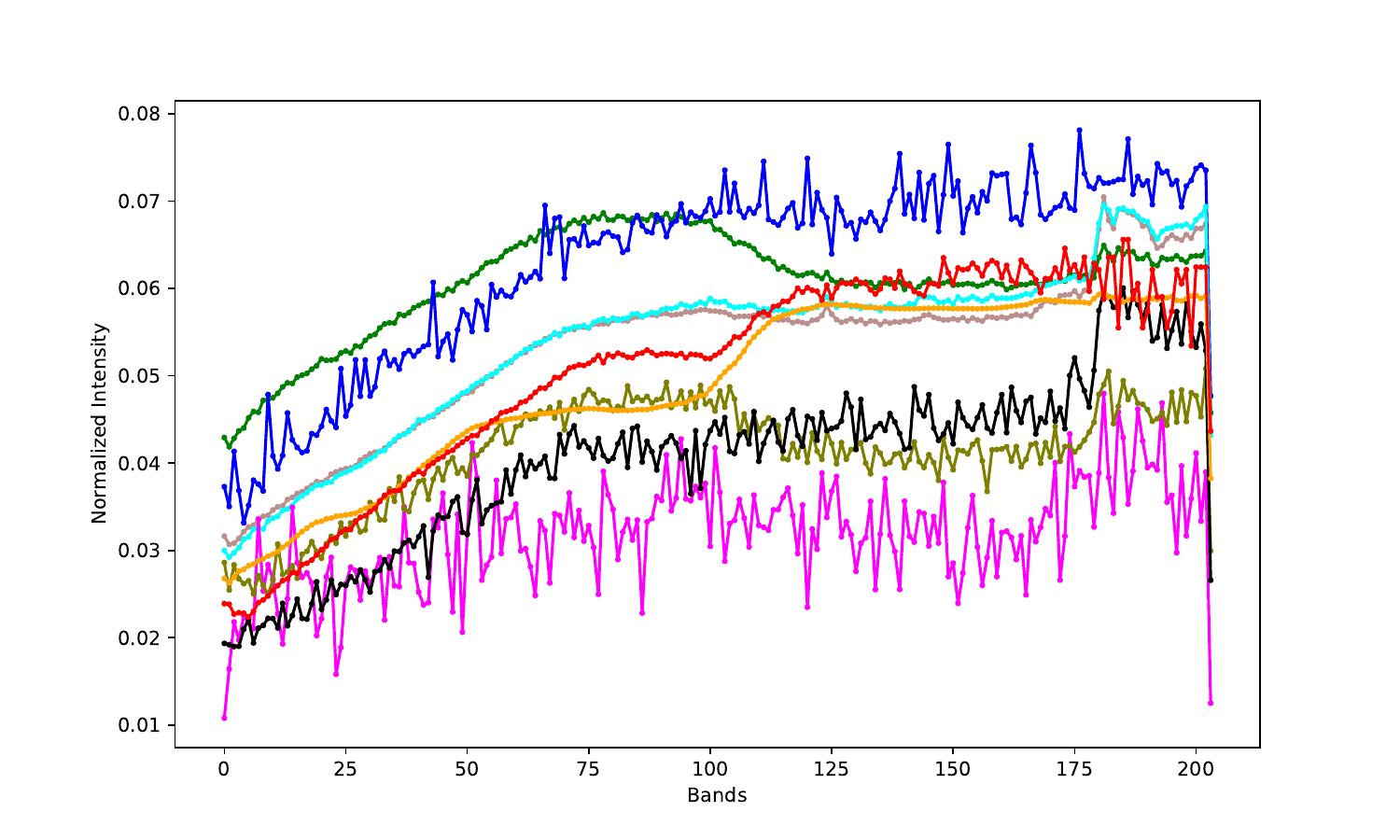}
        \caption{LIB-HSI}
        \label{fig:libhsi1}
    \end{subfigure}
    
    % Row 2
    \begin{subfigure}[b]{0.33\textwidth}
        \includegraphics[width=0.95\columnwidth]{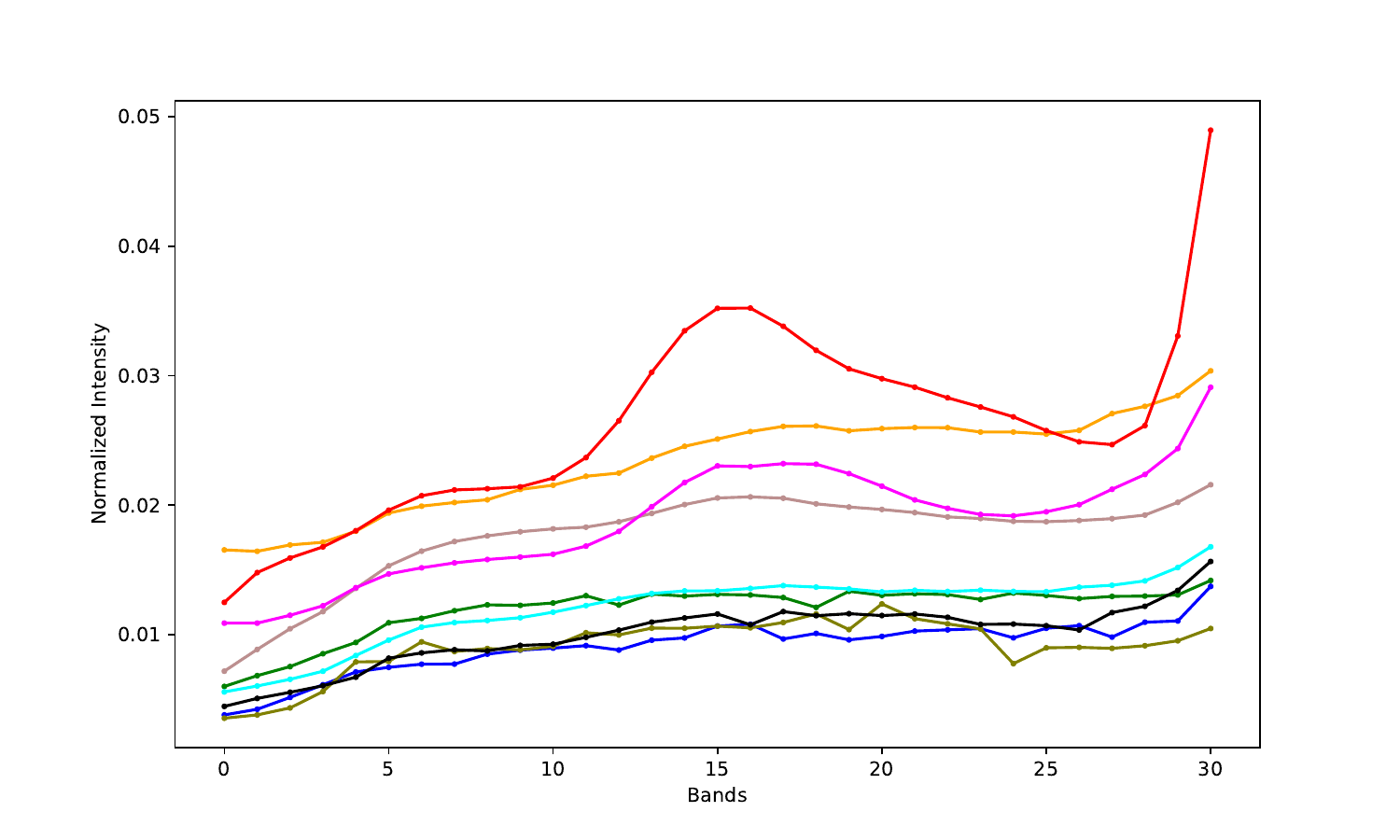}
        % \vspace{-1mm}
        \caption{NTIRE-2022}
        \label{fig:arad2}
    \end{subfigure}
    \begin{subfigure}[b]{0.33\textwidth}
        \includegraphics[width=0.95\columnwidth]{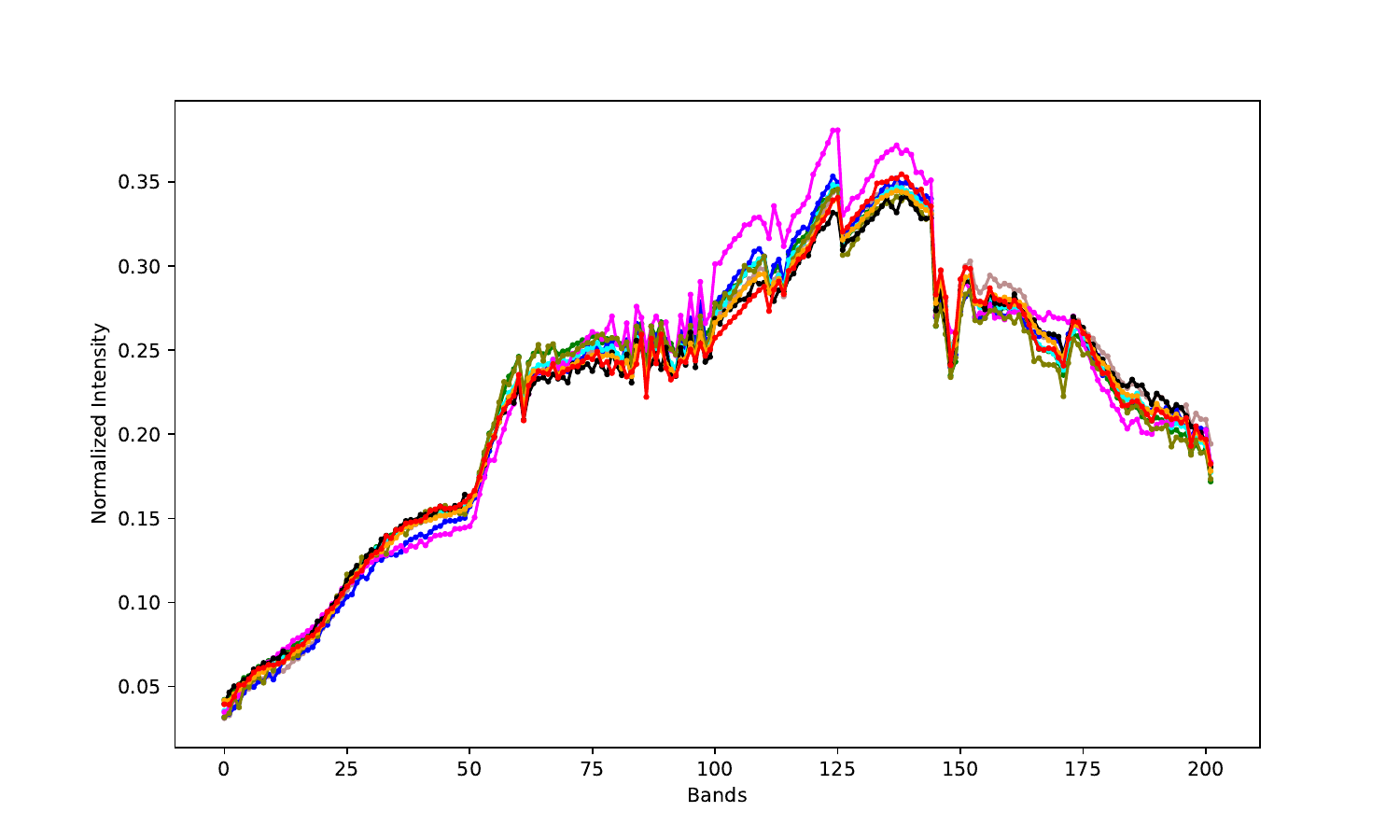}
        \caption{HySpecNet-11k}
        \label{fig:hyspec2}
    \end{subfigure}
    \begin{subfigure}[b]{0.33\textwidth}
        \includegraphics[width=0.95\columnwidth]{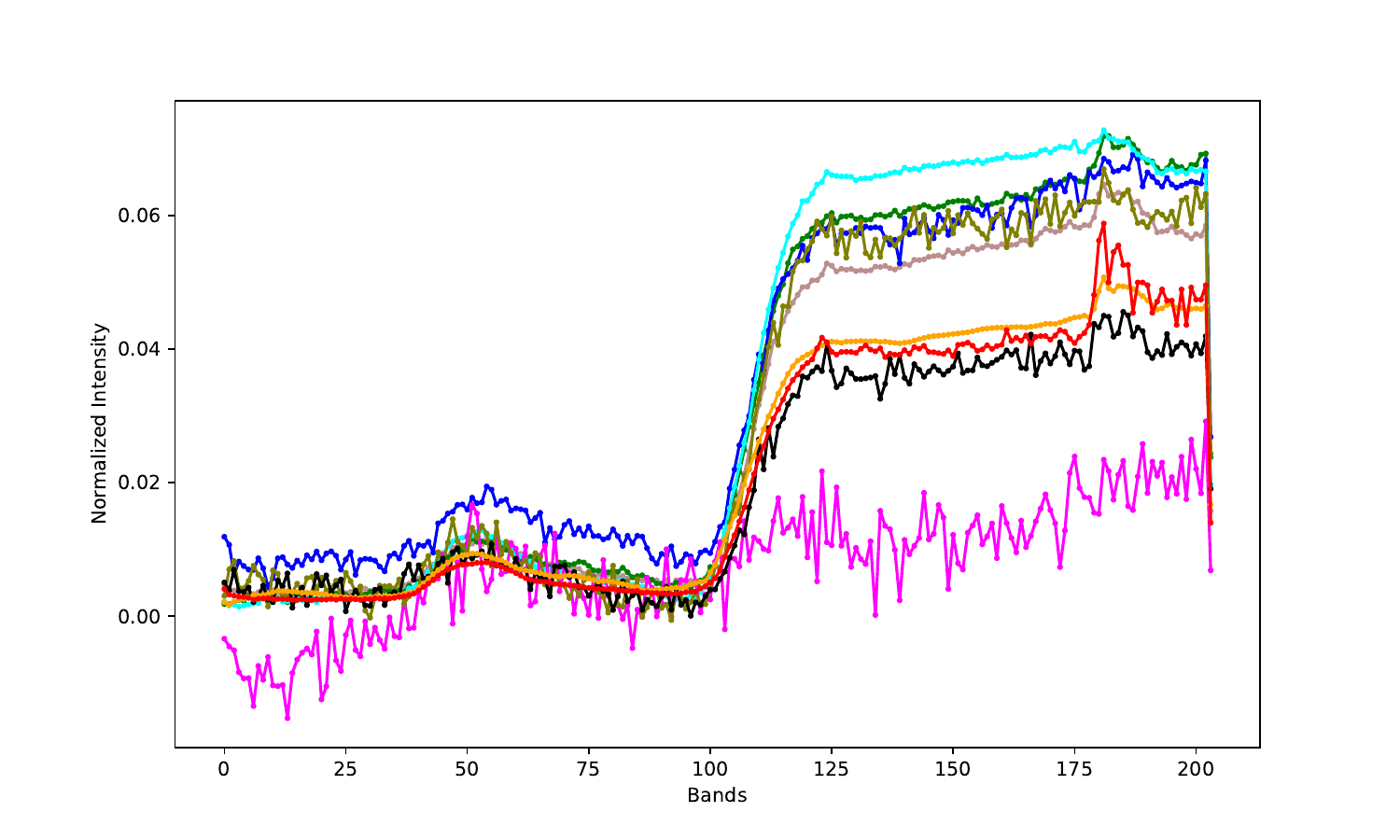}
        \caption{LIB-HSI}
        \label{fig:libhsi2}
    \end{subfigure}
    % Adding negative space before the legend
    \vspace{-2mm}
    \begin{subfigure}[b]{\textwidth}
        \includegraphics[width=\textwidth]{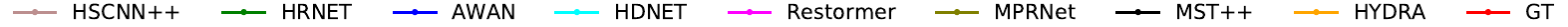} % Update the path to your legend PDF
        % No caption as per requirement
    \end{subfigure}
    \vspace{-5mm}
    \caption{Visualisation of 6 random pixel reconstructions at inference from selected test datasets and approaches}
    \label{fig:pixelRecon}
\vspace{-4mm}
\end{figure*}

We visualised error rates with MRAE heat-maps and input test RGB images (Fig.\ref{fig:heatmaps}); blue marks lower errors, red higher. HYDRA’s superiority is evident in NTIRE-2022, with darker blue hues outshining other models. This stems from our Teacher-Student approach, which constrains the Student’s predictions within the latent space, reducing outliers compared to reconstructing the full high-dimensional spectra directly. Models like Restormer, lacking such guidance, are more prone to errors. In the larger HyspecNet-11k dataset (Fig. \ref{fig:heatmapHyspec}), HYDRA excels in land shade reconstruction, benefiting from the increased dataset size.
The LIB-HSI dataset exhibits artifacts and noise, analyzed further in our pixel reconstruction study (Fig. \ref{fig:pixelRecon}). We compare random pixel spectra from reconstructed HSIs to ground truth (GT in red) across all datasets. Unlike a single global metric, these plots provide a visual analysis of spectral shapes across the visible-to-infrared range. This approach lets us assess how well each model captures the structure of the spectrum without being biased by aggregate accuracy values—i.e., a model might score better on a global error metric yet fail to replicate the correct overall shape. By visually inspecting these spectra, we gain a more nuanced understanding of each model’s strengths and weaknesses in reconstructing hyperspectral data.
% Fig. \ref{fig:pixelRecon} shows NTIRE-2022 yields the best reconstructions, aided by its shallower channel depth. Some discrepancies arise in absolute intensities, but attention-based models generally align closely with GT (Figs. \ref{fig:arad1}, \ref{fig:hyspec1}, \ref{fig:libhsi1}). Further analysis (Figs. \ref{fig:arad2}, \ref{fig:hyspec2}, \ref{fig:libhsi2}) demonstrates HYDRA’s consistently superior performance.
As seen in Fig. \ref{fig:pixelRecon}, the test NTIRE-2022 images show the best reconstruction quality, due to their shallower channel depth. Although the general spectral shapes are well-predicted, some discrepancies in absolute intensity levels occur. Figures \ref{fig:arad1}, \ref{fig:hyspec1}, and \ref{fig:libhsi1} illustrate that models with attention more closely match the ground truth. Further analysis in figures \ref{fig:arad2}, \ref{fig:hyspec2}, and \ref{fig:libhsi2} highlights HYDRA's superior accuracy over other models, with more consistent reconstructions across all datasets.

\subsection{Ablation Study}
Below we analyse how different latent sizes and training stages affect HYDRA's performance (see table \ref{tab:ablation1}). The three training stages are: Teacher-only, Teacher-Student, and Refinement. The Teacher-only column encodes and decodes test HSIs, illustrating the model’s upper bound in stages 2 and 3, but does not address the SR task. HYDRA is sensitive to latent size changes: size 6 is optimal for NTIRE-2022, while 17 is best for HySpecNet-11k and LIB-HSI. This indicates the need to balance representation capacity with minimal latent size for accurate RGB-to-HSI mapping.
Further comparison of three training variations—Stage 1+3 only, Student-only, and the full Three-Stage training is detailed in Table \ref{tab:ablation2}. The comprehensive Three-Stage training approach surpasses the alternatives, underscoring its critical role in enhancing the HYDRA model’s performance across all tested datasets.

\begin{table}[tp]
  \centering
  \caption{Ablation study for HYDRA with different latent-channel sizes and training stages.  
           The Stage 1 column shows the Teacher model’s HSI auto-encoding performance (upper bound for Stages 2–3).}
  \vspace{-2mm}
  \resizebox{\linewidth}{!}{%
    \begin{tabular}{|c|c|ccc|ccc|ccc|}
      \hline
      \textbf{Dataset} &
      \textbf{Latent} &
      \multicolumn{3}{c|}{\textbf{Stage 1\,(Teacher\*)}} &
      \multicolumn{3}{c|}{\textbf{Stage 2\,(Teach.–Stud.)}} &
      \multicolumn{3}{c|}{\textbf{Stage 3\,(Refinement)}} \\ \cline{3-11}
      & \textbf{ Size} &
      MRAE & RMSE & PSNR &
      MRAE & RMSE & PSNR &
      MRAE & RMSE & PSNR \\ \hline

      % ------- NTIRE-2022 -------
      \multirow{3}{*}{NTIRE-2022} 
        & 4  & 0.0419 & 0.0057 & 46.51 & 0.2412 & 0.0328 & 31.89 & 0.2207 & 0.0386 & 32.53 \\
        & 6  & 0.0198 & 0.0027 & 53.17 & 0.1772 & 0.0250 & 33.97 & \textbf{0.1556} & \textbf{0.0221} & \textbf{34.83} \\
        & 8  & 0.0125 & 0.0017 & 56.70 & 0.2164 & 0.0252 & 33.77 & \underline{0.1723} & \underline{0.0245} & \underline{34.02} \\ \hline

      % ------- HySpecNet-11k -------
      \multirow{3}{*}{HySpecNet-11k}
        & 13 & 0.1011 & 0.0047 & 47.35 & 0.1701 & 0.0178 & 37.11 & 0.1658 & 0.0168 & 37.25 \\
        & 17 & 0.0756 & 0.0032 & 50.90 & 0.1621 & 0.0166 & 37.38 & \textbf{0.1563} & \textbf{0.0160} & \textbf{37.76} \\
        & 34 & 0.0683 & 0.0025 & 52.95 & 0.1685 & 0.0171 & 37.32 & \underline{0.1597} & \underline{0.0161} & \underline{37.64} \\ \hline

      % ------- LIB-HSI -------
      \multirow{3}{*}{LIB-HSI}
        & 13 & 0.0228 & 0.0012 & 59.29 & 0.4373 & 0.0139 & 38.55 & 0.4211 & \underline{0.0120} & 39.64 \\
        & 17 & 0.0187 & 0.0010 & 60.49 & 0.3171 & 0.0110 & 41.91 & \textbf{0.3004} & \textbf{0.0091} & \textbf{42.24} \\
        & 34 & 0.0143 & 0.0008 & 62.52 & 0.3278 & 0.0141 & 40.18 & \underline{0.3201} & 0.0125 & \underline{41.84} \\ \hline
    \end{tabular}%
  }
  \label{tab:ablation1}
  \vspace{-5mm}
\end{table}

% \begin{table}[ht]
% \vspace{-4mm}
% \caption{17 Channel latent space comparison of refinement only, Student only, and three-stage training}
% \centering
% \vspace{-2mm}
% \resizebox{\columnwidth}{!}{%
% \begin{tabular}{|c|ccc|ccc|ccc|}
% \hline
% \multirow{2}{*}{Dataset} &
%   \multicolumn{3}{c|}{Stage 1+3 only Training} &
%   \multicolumn{3}{c|}{Student only Training} &
%   \multicolumn{3}{c|}{Three-Stage Training} \\ \cline{2-10} 
%  &
%   \multicolumn{1}{c|}{MRAE} &
%   \multicolumn{1}{c|}{RMSE} &
%   PSNR &
%   \multicolumn{1}{c|}{MRAE} &
%   \multicolumn{1}{c|}{RMSE} &
%   PSNR &
%   \multicolumn{1}{c|}{MRAE} &
%   \multicolumn{1}{c|}{RMSE} &
%   PSNR \\ \hline
% NTIRE-2022 &
%   \multicolumn{1}{c|}{0.1689} &
%   \multicolumn{1}{c|}{0.0273} &
%   32.92 &
%   \multicolumn{1}{c|}{0.1833} &
%   \multicolumn{1}{c|}{0.0274} &
%   33.40 &
%   \multicolumn{1}{c|}{0.1556} &
%   \multicolumn{1}{c|}{0.0221} &
%   34.83 \\ \hline
% HySpecNet-11k &
%   \multicolumn{1}{c|}{0.2689} &
%   \multicolumn{1}{c|}{0.0215} &
%   35.11 &
%   \multicolumn{1}{c|}{0.296} &
%   \multicolumn{1}{c|}{0.0187} &
%   36.04 &
%   \multicolumn{1}{c|}{0.1563} &
%   \multicolumn{1}{c|}{0.0160} &
%   37.76 \\ \hline
% LIB-HSI &
%   \multicolumn{1}{c|}{0.882} &
%   \multicolumn{1}{c|}{0.02294} &
%   33.744 &
%   \multicolumn{1}{c|}{0.3905} &
%   \multicolumn{1}{c|}{0.0131} &
%   38.67 &
%   \multicolumn{1}{c|}{0.3004} &
%   \multicolumn{1}{c|}{0.00905} &
%   42.24 \\ \hline
% \end{tabular}%
% \label{tab:ablation2}
% }

\begin{table}[tp]
  \centering
  % \vspace{-6mm}
  \caption{17-channel latent-space comparison of refinement only, student only, and three-stage training.}
  \vspace{-2mm}
  \resizebox{\linewidth}{!}{%
    \begin{tabular}{|c|ccc|ccc|ccc|}
      \hline
      \textbf{Dataset} &
        \multicolumn{3}{c|}{\textbf{Stage 1\,+\,3}} &
        \multicolumn{3}{c|}{\textbf{Student Only}} &
        \multicolumn{3}{c|}{\textbf{Three-Stage}} \\ \cline{2-10}
      & MRAE & RMSE & PSNR &
        MRAE & RMSE & PSNR &
        MRAE & RMSE & PSNR \\ \hline
      NTIRE-2022    & \underline{0.1689} & 0.0273 & 32.92 & 0.1833 & 0.0274 & 33.40 & \textbf{0.1556} & \textbf{0.0221} & \textbf{34.83} \\
      HySpecNet-11k & 0.2689 & 0.0215 & 35.11 & \underline{0.2960} & \underline{0.0187} & \underline{36.04} & \textbf{0.1563} & \textbf{0.0160} & \textbf{37.76} \\
      LIB-HSI       & 0.8820 & 0.0229 & 33.74 & \underline{0.3905} & \underline{0.0131} & \underline{38.67} & \textbf{0.3004} & \textbf{0.0091} & \textbf{42.24} \\
      \hline
    \end{tabular}%
  }
  \label{tab:ablation2}
  \vspace{-5mm}
\end{table}

% \vspace{-2mm}
% \vspace{-5mm}
% \end{table}

%% file: sec/5_conclusion.tex
\section{Conclusion}
% \vspace{-2mm}
% In this study, we introduce HYDRA, a groundbreaking approach in the field of spectral reconstruction. 
% HYDRA combines the strengths of knowledge distillation, Teacher-Student architectures, and cross-modal learning to surpass current SOTA models in spectral reconstruction tasks. 
% Beyond its superior accuracy on primary spectral reconstruction benchmarks, HYDRA demonstrates exceptional computational efficiency, particularly with increasing hyperspectral channel size, as evidenced by our quantitative results. 
% Crucially, it stands as a particularly effective spectral reconstruction in various channel-depth scenarios, marking it as a pioneering technique for modern HSI applications. 
% For future work, it would be interesting to explore the application of HYDRA in related reconstruction tasks, such as denoising and super-resolution, to potentially improve data efficiency, or to deal with the learning bottleneck that modern transformer systems experience on high-dimensional data.
We introduce HYDRA, a novel spectral reconstruction approach that leverages knowledge distillation, Teacher-Student architectures, and cross-modal learning to outperform existing SOTA models. 
HYDRA not only achieves superior accuracy on primary benchmarks but also demonstrates exceptional computational efficiency across varying hyperspectral channel sizes. 
It excels in diverse channel-depth scenarios, positioning it as a novel technique for future HSI applications. 
% Future work could explore HYDRA's potential in denoising, super-resolution, and addressing the learning bottlenecks of modern transformers in high-dimensional data.